\relax
\documentclass[runningheads]{llncs}

\usepackage{times}  
\usepackage{helvet}  
\usepackage{courier}  
\usepackage{url}  
\usepackage{graphicx}  
\usepackage{algpseudocode}
\usepackage{algorithm}
\usepackage{subcaption}
\begin{document}
%

%

\title{A Computational Approach to Measuring \\the Semantic Divergence of Cognates} 

\author{Ana-Sabina Uban\inst{1,3} \and Alina Ciobanu\inst{1,2} \and Liviu P. Dinu\inst{1,2}}
%
%
\institute{Faculty of Mathematics and Computer Science \and
Human Language Technologies Research Center \and 
Data Science Center \\ University of Bucharest \\
{\tt ana.uban@gmail.com}, {\tt alina.ciobanu@my.fmi.unibuc.ro}, {\tt liviu.p.dinu@gmail.com}}

%
\maketitle

\begin{abstract}
   Meaning is the foundation stone of intercultural communication. Languages are continuously changing, and words shift their meanings for various reasons. Semantic divergence in related languages is a key concern of historical linguistics. In this paper we investigate semantic divergence across languages by measuring the semantic similarity of cognate sets in multiple languages. The method that we propose is based on cross-lingual word embeddings. In this paper we implement and evaluate our method on English and five Romance languages, but it can be extended easily to any language pair, requiring only large monolingual corpora for the involved languages and a small bilingual dictionary for the pair. This language-agnostic method facilitates a quantitative analysis of cognates divergence -- by computing degrees of semantic similarity between cognate pairs -- and provides insights for identifying false friends. As a second contribution, we formulate a straightforward method for detecting false friends, and introduce the notion of "soft false friend" and "hard false friend", as well as a measure of the degree of "falseness" of a false friends pair. Additionally, we propose an algorithm that can output suggestions for correcting false friends, which could result in a very helpful tool for language learning or translation.
\end{abstract}

\section{Introduction}

Semantic change -- that is, change in the meaning of individual words \cite{campbell_1998} -- is a continuous, inevitable process stemming from numerous reasons and influenced by various factors. Words are continuously changing, with new senses emerging all the time. \cite{campbell_1998} presents no less than 11 types of semantic change, that are generally classified in two wide categories: narrowing and widening.
Most linguists found structural and psychological factors to be the main cause of semantic change, but the evolution of technology and cultural and social changes are not to be omitted.

Measuring semantic divergence across languages can be useful in theoretical and historical linguistics -- being central to models of language and cultural evolution -- but also in downstream applications relying on cognates, such as machine translation.

\begin{figure}[ht]
\center
\includegraphics[width=200pt]{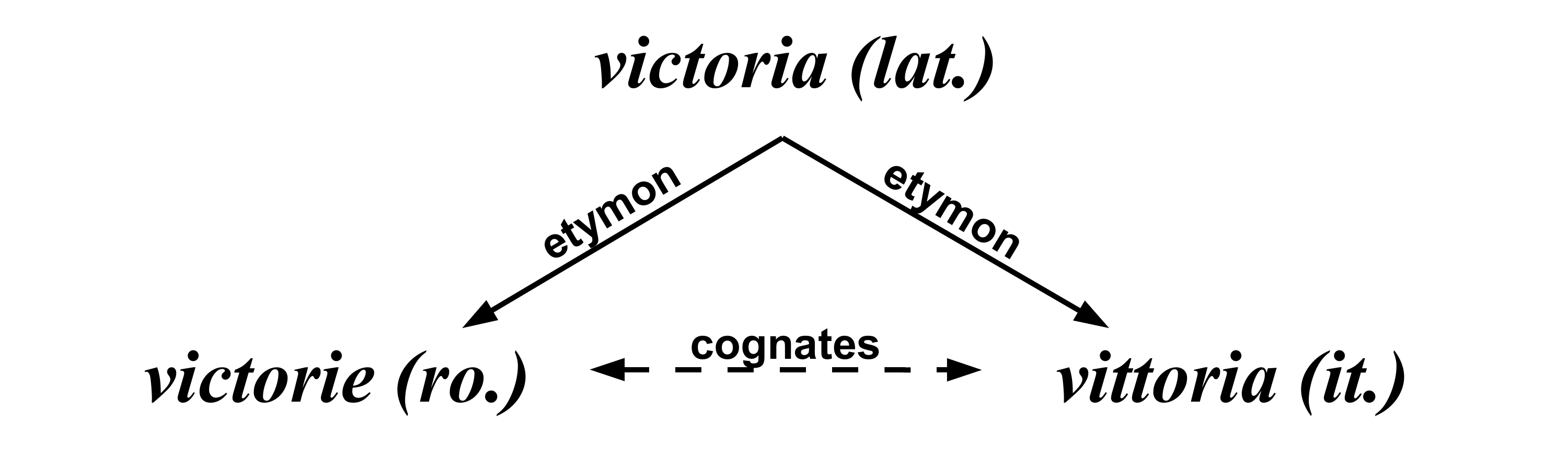}
\caption{\label{fig:cognates}Example of cognates and their common ancestor}
\end{figure}

\textbf{Cognates} are words in sister languages (languages descending from a common anscestor) with a common proto-word. For example, the Romanian word \emph{victorie} and the Italian word \emph{vittoria} are cognates, as they both descend from the Latin word \emph{victoria} (meaning \emph{victory}) -- see Figure \ref{fig:cognates}. In most cases, cognates have preserved similar meanings across languages, but there are also exceptions. These are called deceptive cognates or, more commonly, false friends. Here we use the definition of cognates that refers to words with similar appearance and some common etymology, and use "true cognates" to refer to cognates which also have a common meaning, and "deceptive cognates" or "false friends" to refer to cognate pairs which do not have the same meaning (anymore). The most common way cognates have diverged is by changing their meaning. For many cognate pairs, however, the changes can be more subtle, relating to the feeling attached to a word, or its conotations. This can make false friends even more delicate to distinguish from true cognates.

Cognate word pairs can help students when learning a second language and contributes to the
expansion of their vocabularies. False friends, however, from the more obvious differences in meaning to the more subtle, have the opposite effect, and can be confusing for language learners and make the correct use of language more difficult.
Cognate sets have also been used in a number applications in natural language processing, including for example machine translation \cite{kondrak2003cognates}. These applications rely on properly distinguishing between true cognates and false friends.


\subsection{Related work}


Cross-lingual semantic word similarity consists in identifying words that refer to similar semantic concepts and convey similar meanings across languages \cite{vulic_and_moens_2}. Some of the most popular approaches rely on probabilistic models \cite{vulic_and_moens} and cross-lingual word embeddings \cite{soegard_et_al}.

A comprehensive list of cognates and false friends for every language pair is difficult to find or manually build - this is why applications have to rely on automatically identifying them. There have been a number of previous studies attempting to automatically extract pairs of true cognates and false friends from corpora or from dictionaries. Most methods are based either on ortographic and phonetic similarity, or require large parallel corpora or dictionaries \cite{inkpen2005automatic,st2017identifying,nakov2009unsupervised,chen2016false}. We propose a corpus-based approach that is capable of covering the vast majority of the vocabulary for a large number of languages, while at the same time requiring minimal human effort in terms of manually evaluating word pairs similarity or building lexicons, requiring only large monolingual corpora.

In this paper, we make use of cross-lingual word embeddings in order to distinguish between true cognates and false friends. There have been few previous studies using word embeddings for the detection of false friends or cognate words, usually using simple methods on only one or two pairs of languages \cite{castro2018high,torres2011using}.



\begin{figure*}[!ht]
    \centering
    \begin{subfigure}{0.20\textwidth}
        \includegraphics[width=\linewidth]{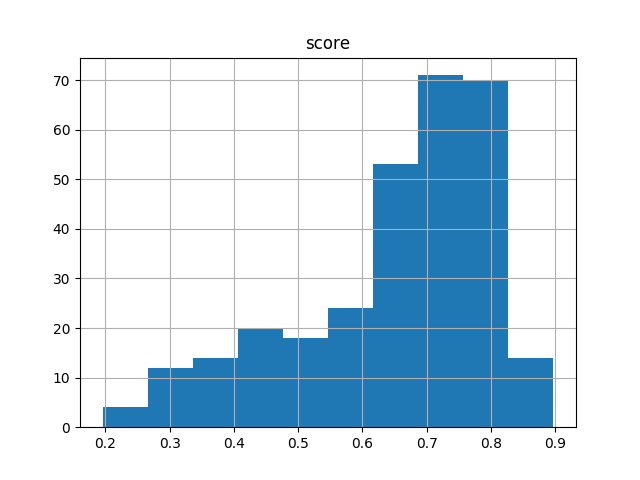}
        \caption{Es-Fr}
    \end{subfigure}
    \begin{subfigure}{0.20\textwidth}
        \includegraphics[width=\linewidth]{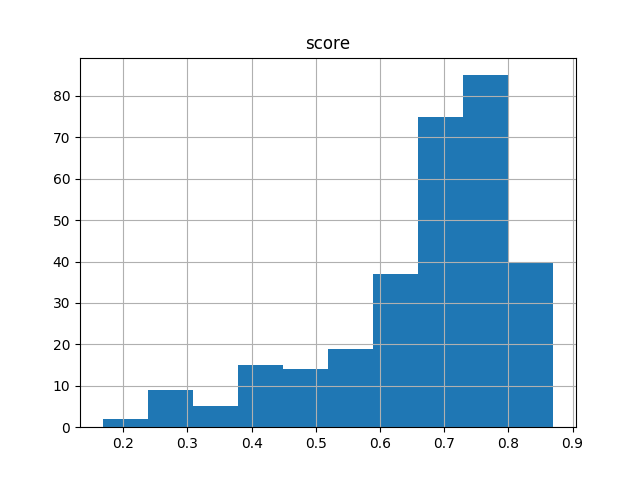}
        \caption{Es-It}
    \end{subfigure}
    \begin{subfigure}{0.20\textwidth}
        \includegraphics[width=\linewidth]{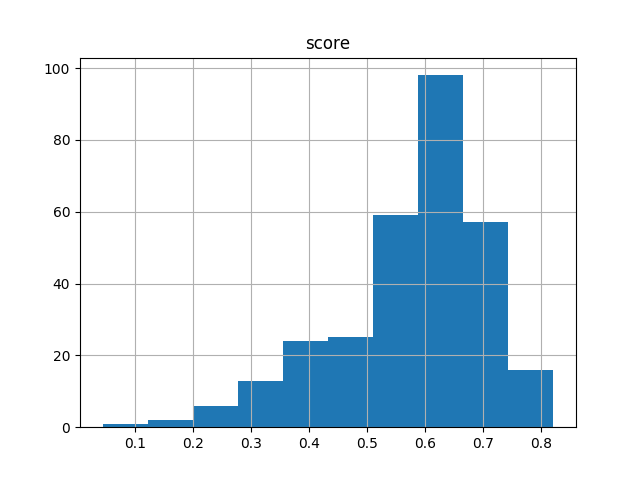}
        \caption{Es-Ro}
    \end{subfigure}
    \begin{subfigure}{0.20\textwidth}
        \includegraphics[width=\linewidth]{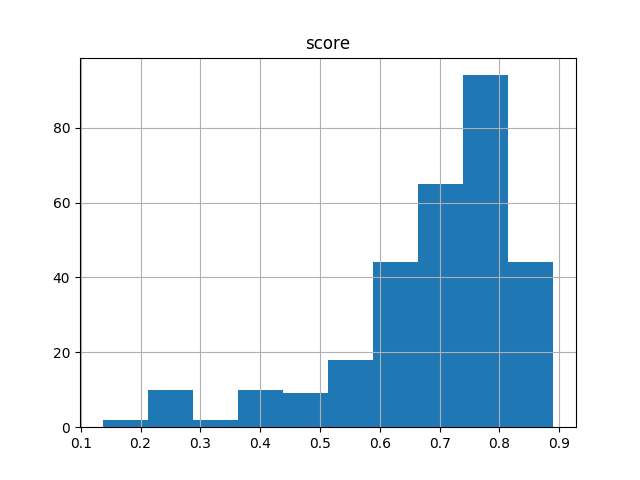}
        \caption{Es-Pt}
    \end{subfigure}
    \begin{subfigure}{0.20\textwidth}
        \includegraphics[width=\linewidth]{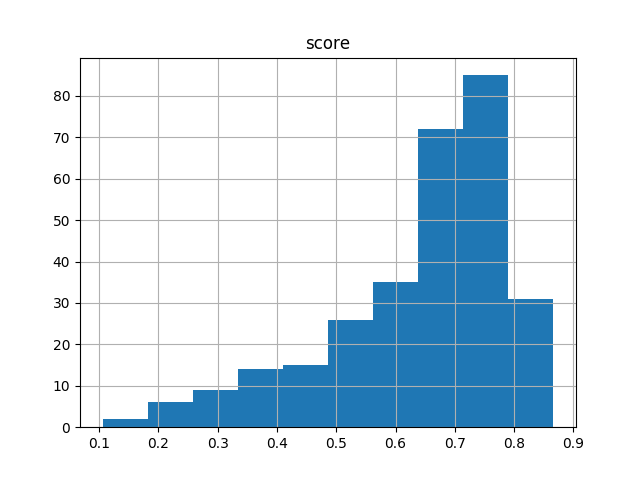}
        \caption{Fr-It}
    \end{subfigure}
    \begin{subfigure}{0.20\textwidth}
        \includegraphics[width=\linewidth]{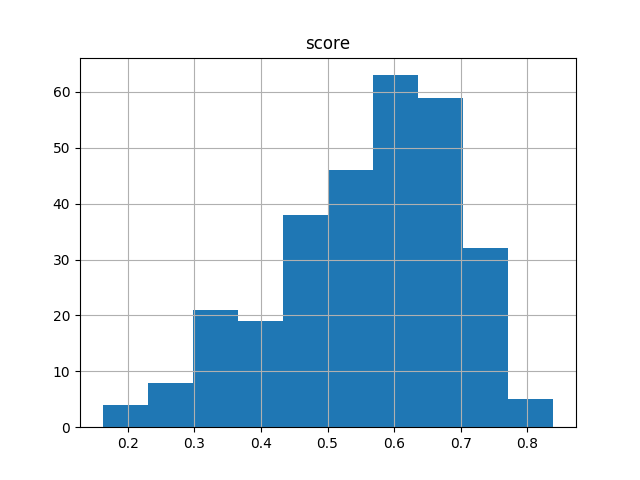}
        \caption{Fr-Ro}
    \end{subfigure}
    \begin{subfigure}{0.20\textwidth}
        \includegraphics[width=\linewidth]{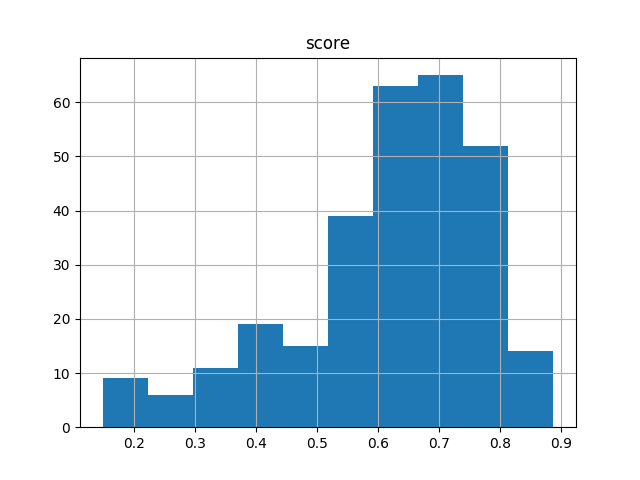}
        \caption{Fr-Pt}
    \end{subfigure}
    \begin{subfigure}{0.23\textwidth}
        \includegraphics[width=\linewidth]{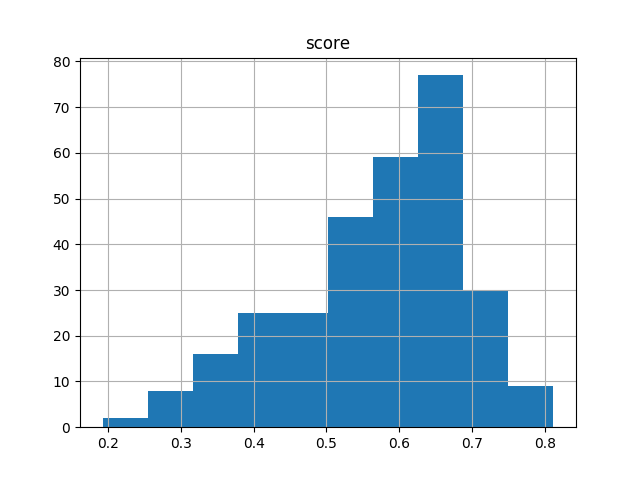}
        \caption{It-Ro}
    \end{subfigure}
    \begin{subfigure}{0.20\textwidth}
        \includegraphics[width=\linewidth]{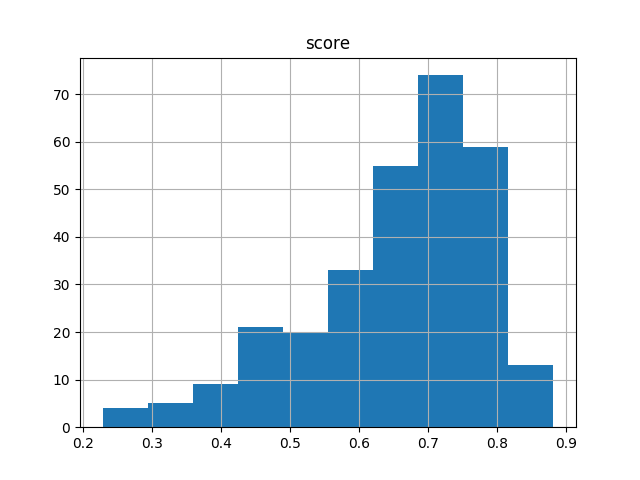}
        \caption{It-Pt}
    \end{subfigure}
    \begin{subfigure}{0.20\textwidth}
        \includegraphics[width=\linewidth]{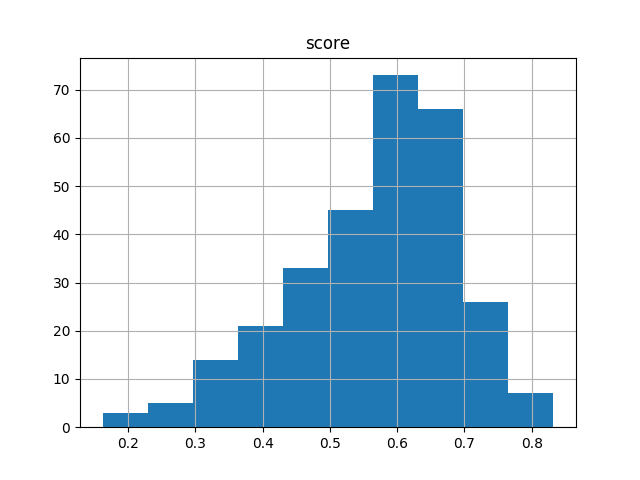}
        \caption{Ro-Pt}
    \end{subfigure}
    \begin{subfigure}{0.20\textwidth}
        \includegraphics[width=\linewidth]{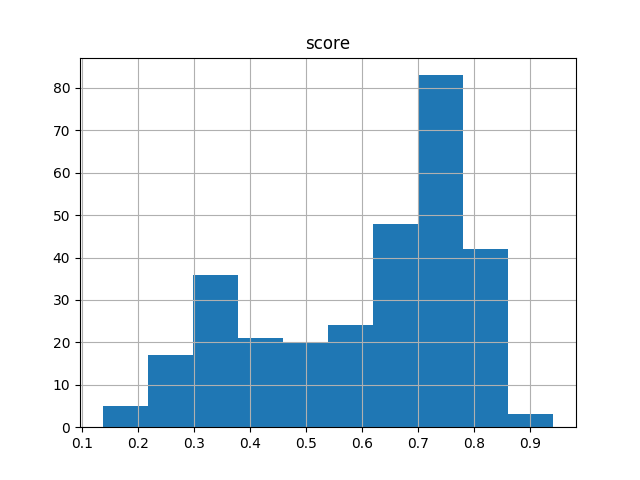}
        \caption{En-Fr}
    \end{subfigure}
    \begin{subfigure}{0.20\textwidth}
        \includegraphics[width=\linewidth]{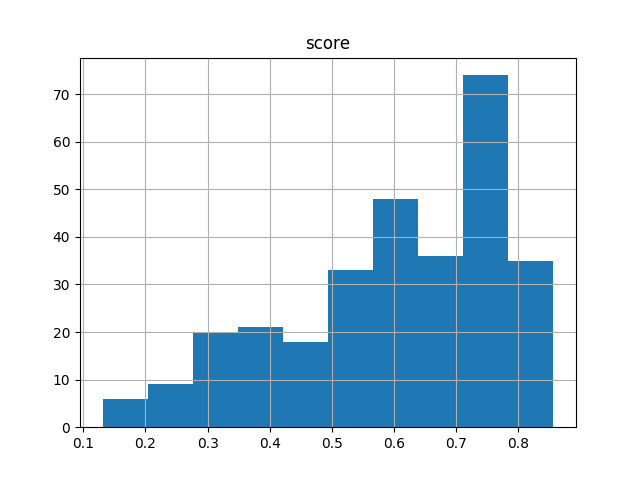}
        \caption{En-It}
    \end{subfigure}
    \begin{subfigure}{0.20\textwidth}
        \includegraphics[width=\linewidth]{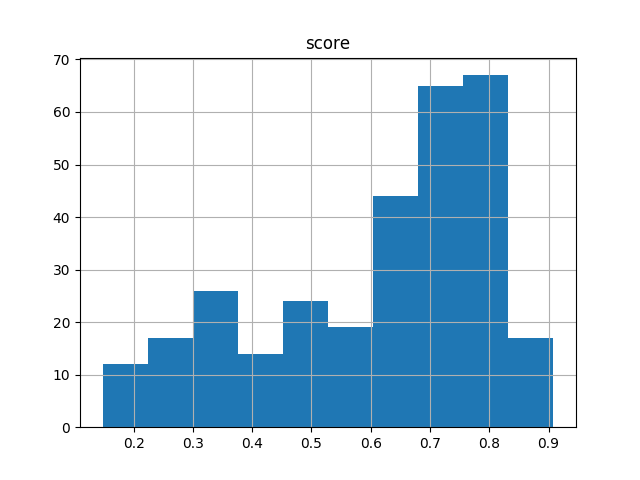}
        \caption{En-Es}
    \end{subfigure}
    \begin{subfigure}{0.23\textwidth}
        \includegraphics[width=\linewidth]{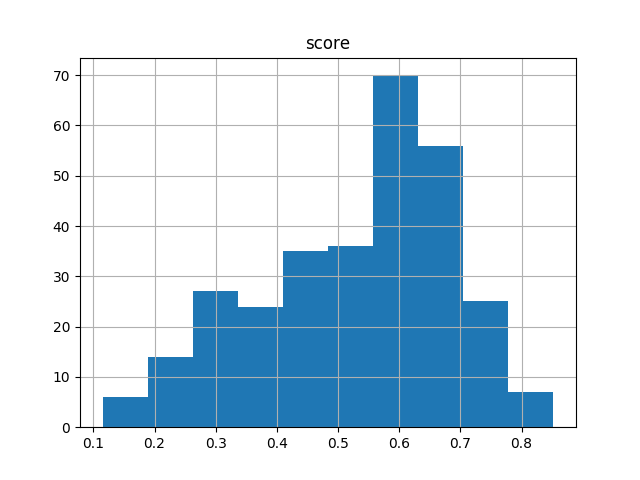}
        \caption{En-Ro}
    \end{subfigure}
    \begin{subfigure}{0.20\textwidth}
        \includegraphics[width=\linewidth]{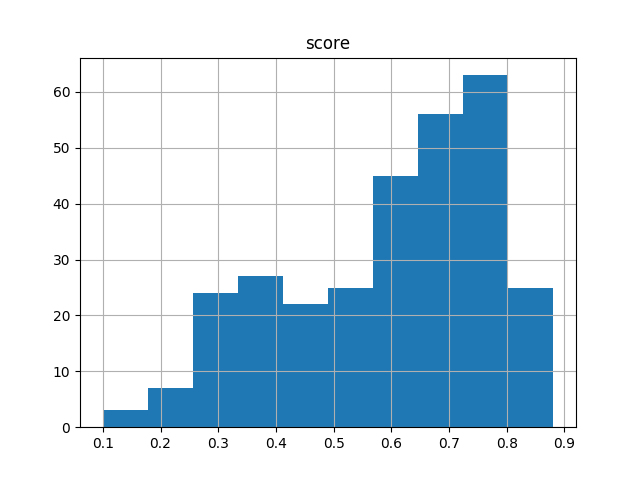}
        \caption{En-Pt}
    \end{subfigure}
    \begin{subfigure}{0.20\textwidth}
        \includegraphics[width=\linewidth]{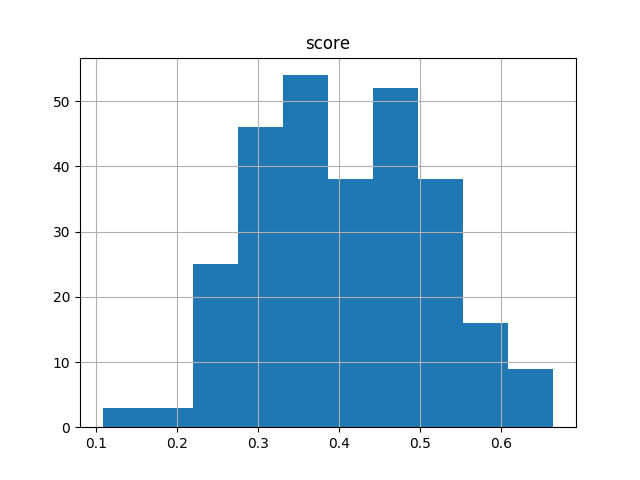}
        \caption{En-La}
    \end{subfigure}
    \begin{subfigure}{0.20\textwidth}
        \includegraphics[width=\linewidth]{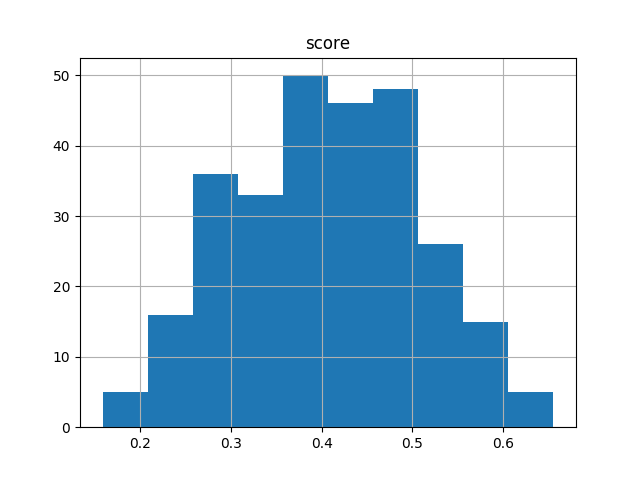}
        \caption{Fr-La}
    \end{subfigure}
    \begin{subfigure}{0.20\textwidth}
        \includegraphics[width=\linewidth]{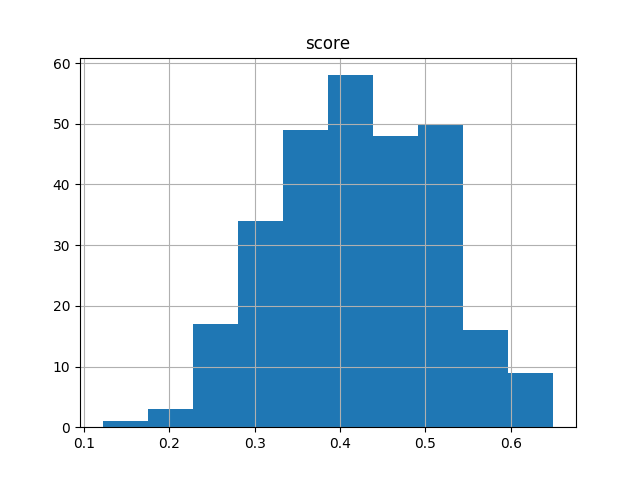}
        \caption{Es-La}
    \end{subfigure}
    \begin{subfigure}{0.20\textwidth}
        \includegraphics[width=\linewidth]{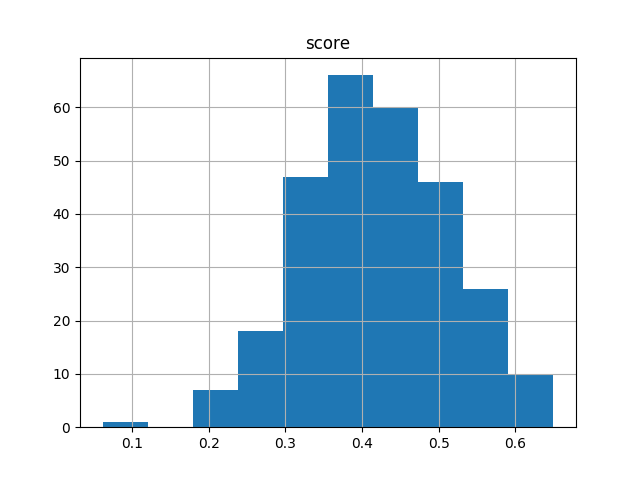}
        \caption{It-La}
    \end{subfigure}
    \begin{subfigure}{0.20\textwidth}
        \includegraphics[width=\linewidth]{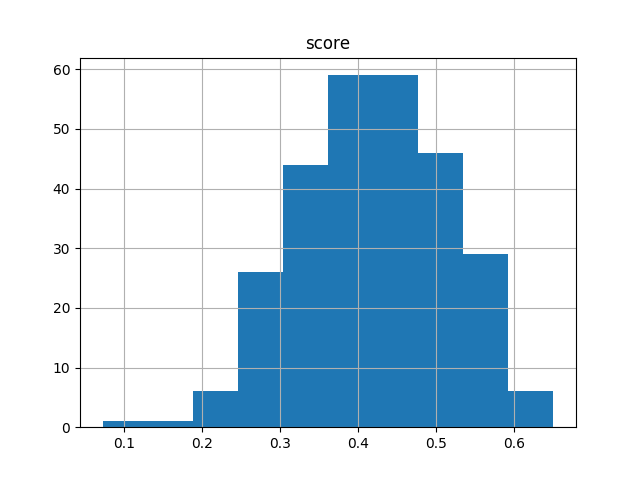}
        \caption{Pt-La}
    \end{subfigure}
    \begin{subfigure}{0.20\textwidth}
        \includegraphics[width=\linewidth]{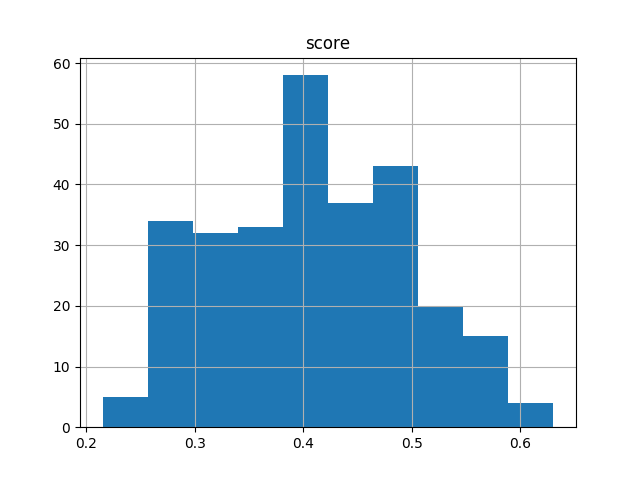}
        \caption{Ro-La}
    \end{subfigure}
    \caption{\label{fig:exp1}Distributions of cross-language similarity scores between cognates.}
\end{figure*}

\subsection{Contributions}
The contributions of our paper are twofold: firstly, we propose a method for quantifying the semantic divergence of languages; secondly, we provide a framework for detecting and correcting false friends, based on the observation that these are usually deceptive cognate pairs: pairs of words that once had a common meaning, but whose meaning has since diverged.

We propose a method for measuring the semantic divergence of sister languages based on cross-lingual word embeddings. We report empirical results on five Romance languages: Romanian, French, Italian, Spanish and Portuguese.
For a deeper insight into the matter, we also compute and investigate the semantic similarity betweeen modern Romance languages and Latin. We finally introduce English into the mix, to analyze the behavior of a more remote language, where words deriving from Latin are mostly borrowings.

Further, we make use of cross-lingual word embeddings in order to distinguish between true cognates and false friends. There have been few previous studies using word embeddings for the detection of false friends or cognate words, usually using simple methods on only one or two pairs of languages \cite{castro2018high,torres2011using}.

Our chosen method of leveraging word embeddings extends naturally to another application related to this task which, to our knowledge, has not been explored so far in research: false friend correction. We propose a straightforward method for solving this task of automatically suggesting a replacement when a false friend is incorrectly used in a translation. Especially for language learners, solving this problem could result in a very useful tool to help them use language correctly.

\section{The Method}

\subsection{Cross-lingual Word Embeddings}

Word embeddings are vectorial representations of words in a continuous space, built by training a model to predict the occurence of a given word in a text corpus given its context. Based on the distributional hypothesis stating that similar words occur in similar contexts, these vectorial representations can be seen as semantic representations of words and can be used to compute semantic similarity between word pairs (representations of words with similar meanings are expected to be close together in the embeddings space).

To compute the semantic divergence of cognates across sister languages, as well as identify pairs of false cognates (pairs of cognates with high semantic distance), which by definition are pairs of words in two different languages, we need to obtain a multilingual semantic space, which is shared between the cognates. Having the representations of both cognates in the same semantic space, we can then compute the semantic distance between them using their vectorial representations in this space.

We use word embeddings computed using the FastText algorithm, pre-trained on Wikipedia for the six languages in question. The vectors have dimension 300, and were obtained using the skip-gram model described in \cite{bojanowski_et_al} with default parameters.

The algorithm for measuring the semantic distance between cognates in a pair of languages $ (lang1, lang2) $ consists of the following steps:

\begin{enumerate}
\item Obtain word embeddings for each of the two languages.
\item Obtain a shared embedding space, common to the two languages. This is accomplished using an alignment algorithm, which consists of finding a linear transformation between the two spaces, that on average optimally transforms each vector in one embedding space into a vector in the second embedding space, minimizing the distance between a few seed word pairs (for which it is known that they have the same meaning), based on a small bilingual dictionary. For our purposes, we use the publicly available multilingual alignment matrices that were published in \cite{smith2017offline}.
\item Compute semantic distances for each pair of cognates words in the two languages, using a vectorial distance (we chose cosine distance) on their corresponding vectors in the shared embedding space.
\end{enumerate}

\subsection{Cross-language Semantic Divergence}

We propose a definition of semantic divergence between two languages based on the semantic distances of their cognate word pairs in these embedding spaces. The semantic distance between two languages can then be computed as the average the semantic divergence of each pair of cognates in that language pair.

We use the list of cognates sets in Romance languages proposed by \cite{ciobanu_and_dinu_lrec}. It contains 3,218 complete cognate sets in Romanian, French, Italian, Spanish and Portuguese, along with their Latin common ancestors. The cognate sets are obtained from electronic dictionaries which provide information about the etymology of the words. Two words are considered cognates if they have the same etymon (i.e., if they descend from the same word).

The algorithm described above for computing semantic distance for cognate pairs stands on the assumption that the (shared) embedding spaces are comparable, so that the averaged cosine similarities, as well as the overall distributions of scores that we obtain for each pair of languages can be compared in a meaningful way. For this to be true, at least two conditions need to hold:

\begin{enumerate}
\item The embeddings spaces for each language need to be similarly representative of language, or trained on similar texts - this assumption holds sufficiently in our case, since all embeddings (for all languages) are trained on Wikipedia, which at least contains a similar selection of texts for each language, and at most can be considered comparable corpora.

\item The similarity scores in a certain (shared) embeddings space need to be sampled from a similar distribution. To confirm this assumption, we did a brief experiment looking at the distributions of a random sample of similarity scores across all embeddings spaces, and did find that the distributions for each language pair are similar (in mean and variance). This result was not obvious but also not surprising, since:
\begin{itemize}
    \item The way we create shared embedding spaces is by aligning the embedding space of any language to the English embedding space (which is a common reference to all shared embedding spaces).
    \item The nature of the alignment operation (consisting only of rotations and reflections) guarantees monolingual invariance, as described in these papers: \cite{artetxe2016learning,smith2017offline}.
\end{itemize}
\end{enumerate}

\subsubsection{The Romance Languages}

We compute the cosine similarity between cognates for each pair of modern languages, and between modern languages and Latin as well. We compute an overall score of similarity for a pair of languages as the average similarity for the entire dataset of cognates. The results are reported in Table \ref{table:exp1}.

\begin{table}[!ht]
\begin{center}
\begin{tabular}{l | l l l l l}
\hline
& Fr & It & Pt & Ro & La\\
\hline

Es & 0.67 & 0.69 & 0.70 & 0.58 & 0.41\\
Fr & & 0.66 & 0.64 & 0.56 & 0.40 \\
It & & & 0.66 & 0.57 & 0.41 \\
Pt & & & & 0.57 & 0.41\\
Ro & & & & & 0.40 \\
\hline

\end{tabular}
\end{center}
\caption{\label{table:exp1}Average cross-language similarity between cognates (Romance languages).}
\end{table}

We observe that the highest similarity is obtained between Spanish and Portuguese (0.70), while the lowest are obtained for Latin. From the modern languages, Romanian has, overall, the lowest degrees of similarity to the other Romance languages. A possible explanation for this result is the fact that Romanian developed far from the Romance kernel, being surrounded by Slavic languages. 
In Table \ref{table:exp3} we report, for each pair of languages, the most similar (above the main diagonal) and the most dissimilar (below the main diagonal) cognate pair for Romance languages.

\begin{table*}[!ht]
\begin{small}
\begin{center}
\begin{tabular}{l | l l l l l}
\hline
& Es & Fr & It & Ro & Pt\\
\hline

Es & -- & ocho/huit(0.89) & diez/dieci(0.86) & ocho/opt(0.82) & ocho/oito(0.89) \\
Fr & caisse/casar(0.05) & -- & dix/dieci(0.86) & décembre/decembrie(0.83) & huit/oito(0.88) \\
It & prezzo/prez(0.06) & punto/ponte(0.09) & convincere/convinge(0.75) & convincere/convencer(0.88) \\
Ro & miere/mel(0.09) & face/facteur(0.10) & as/asso(0.11) & -- & opt/oito(0.83) \\
Pt & prez/preço(0.05) & pena/paner(0.09) & preda/prea(0.08) & linho/in(0.05) --\\
\hline

\end{tabular}
\end{center}
\caption{\label{table:exp3}Most similar and most dissimilar cognates}
\end{small}
\end{table*}

\begin{figure}[ht]
\center
\includegraphics[width=0.8\linewidth]{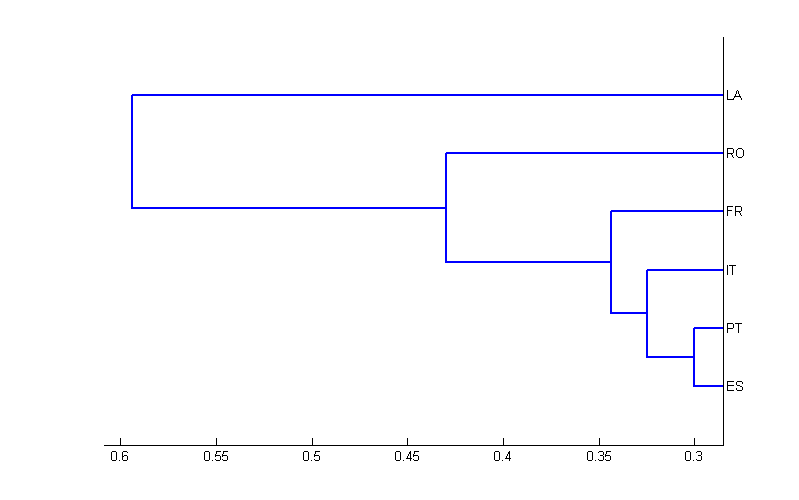}
\caption{\label{fig:dendrogram}Dendrogram of the language clusters}
\end{figure}

The problem that we address in this experiment involves a certain \textit{vagueness of reported values} (also noted by \cite{eger_et_al} in the problem of semantic language classification), as there isn't a gold standard that we can compare our results to. To overcome this drawback, we use the degrees of similarity that we obtained to produce a language clustering (using the UPGMA hierarchical clusering algorithm), and observe that it is similar with the generally accepted tree of languages, and with the clustering tree built on intelligibility degrees by \cite{ciobanu_and_dinu_intelligibility}. The obtained dendrogram is rendered in figure \ref{fig:dendrogram}.

\subsubsection{The Romance Languages vs English}

Further, we introduce English into the mix as well. We run this experiment on a subset of the used dataset, comprising the words that have a cognate in English as well\footnote{Here we \textit{stretch} the definition of \textit{cognates}, as they are generally referring to sister languages. In this case English is not a sister of the Romance languages, and the words with Latin ancestors that entered English are mostly borrowings.}. The subset has 305 complete cognate sets. 

The results are reported in Table \ref{table:exp2}, and the distribution of similarity scores for each pair of languages is rendered in figure \ref{fig:exp1}. We notice that English has 0.40 similarity with Latin, the lowest value (along with French and Romanian), but close to the other languages. Out of the modern Romance languages, Romanian is the most distant from English, with 0.53 similarity.

Another interesting observation relates to the distributions of scores for each language pair, shown in the histograms in \ref{fig:exp1}. While similarity scores between cognates among romance languages usually follow a normal distribution (or another unimodal, more skewed distribution), the distributions of scores for romance languages with English seem to follow a bimodal distribution, pointing to a different semantic evolution for words in English that share a common etymology with a word in a romance language. One possible explanation is that the set of cognates between English and romance languages (which are pairs of languages that are more distantly related) consist of two distinct groups: for example one group of words that were borrowed directly from the romance language to English (which should have more meaning in common), and words that had a more complicated etymological trail between languages (and for which meaning might have diverged more, leading to lower similarity scores). 

\begin{table}[!ht]
\begin{center}
\begin{tabular}{l | l l l l l l}
\hline
& Fr & It & Pt & Ro & En & La\\
\hline

Es & 0.64 & 0.67 & 0.68 & 0.57 & 0.61 & 0.42 \\
Fr & & 0.64 & 0.61 & 0.55 & 0.60 & 0.40 \\
It & & & 0.65 & 0.57 & 0.60 & 0.41 \\
Pt & & & & 0.56 & 0.59 & 0.42 \\
Ro & & & & & 0.53 & 0.40 \\
En & & & & & & 0.40 \\
\hline

\end{tabular}
\end{center}
\caption{\label{table:exp2}Average cross-language similarity between cognates}
\end{table}



\subsection{Detection and Correction of False Friends}

In a second series of experiments, we propose a method for identifying and correcting false friends.
Using the same principles as in the previous experiment, we can use embedding spaces and semantic distances between cognates in order to detect pairs of false friends, which are simply defined as pairs of cognates which do not share the same meaning, or which are not semantically similar \textit{enough}.

This definition is of course ambiguous: there are different degrees of similarity, and as a consequence different potential degrees of \textit{falseness} in a false friend. Based on this observation, we define the notions of \textit{hard false friend} and \textit{soft false friend}.

A \textit{hard false friend} is a pair of cognates for which the meanings of the two words have diverged enough such that they don't have the same meaning anymore, and should not be used interchangibly (as translations of one another). In this category fall most known examples of false friends, such as the French-English cognate pair \textit{attendre} / \textit{attend}: in French, \textit{attendre} has a completely different meaning, which is \textit{to wait}. A different and more subtle type of false friends can result from more minor semantic shifts between the cognates. In such pairs, the meaning of the cognate words may remain roughly the same, but with a difference in nuance or connotation. Such an example is the Romanian-Italian cognate pair \textit{amic} / \textit{amico}. Here, both cognates mean \textit{friend}, but in Italian the conotation is that of a closer friend, whereas the Romanian \textit{amic} denotes a more distant friend, or even aquaintance. A more suitable Romanian translation for \textit{amico} would be \textit{prieten}, while a better translation in Italian for \textit{amic} could be \textit{conoscente}.

Though their meaning is roughly the same, translating one word for the other would be an inaccurate use of the language. These cases are especially difficult to handle by beginner language learners (especially since the cognate pair may appear as valid a translation in multilingual dictionaries) and using them in the wrong contexts is an easy trap to fall into.

Given these considerations, an automatic method for finding the appropriate term to translate a cognate instead of using the false friend would be a useful tool to aid in translation or in language learning.

As a potential solution to this problem, we propose a method that can be used to identify pairs of false friends, to distinguish between the two categories of false friends defined above (\textit{hard false friends} and \textit{soft false friends}), and to provide suggestions for correcting the erroneous usage of a false friend in translation.

\textbf{False friends} can be identified as pairs of cognates with high semantic distance. More specifically, we consider a pair of cognates to be a false friend pair if in the shared semantic space, there exists a word in the second language which is semantically closer to the original word than its cognate in that language (in other words, the cognate is not the optimal translation). The arithmetic difference between the semantic distance between these words and the semantic distance between the cognates will be used as a measure of the \textit{falseness} of the false friend. The word that is found to be closest to the first cognate will be the suggested "correction".

The algorithm can be described as follows:
\begin{algorithm}
\caption{Detection and correction of false friends}
\label{Detection and correction of false friends}
\begin{algorithmic}[1]
    \State Given the cognates pair $(c_1, c_2)$ where $c_1$ is a word in $lang_1$ and $c_2$ is a word in $lang_2$:
    \State Find the word $w_2$ in $lang_2$ such that for any $w_i$ in $lang_2$, $distance(c_2, w_2) < distance(c_2, w_i)$
    \If{$w_2 \neq c_2$}
    \State $(c_1, c_2)$ is a pair of false friends
    \State Degree of falseness $= distance(c_1, w_2) - distance(c_1, c_2)$ \\
    \Return $w_2$ as potential correction
    \EndIf
\end{algorithmic}
\end{algorithm}

We select a few results of the algorithm to show in Table \ref{table:exp3}, containing examples of extracted false friends for the language pair French-Spanish, along with the suggested correction and the computed degree of falseness.

Depending on the application, the measure of \textit{falseness} could be used by choosing a threshold to single out pairs of false friends that are \textit{harder} or \textit{softer}, with a customizable degree of sensitivity to the difference in meaning.

\begin{table}[!ht]
\begin{center}
\begin{tabular}{l l l l}
\hline
FR cognate & ES cognate & Correction & Falseness \\
\hline
prix & prez & premio & 0.67 \\
long & luengo & largo & 0.57 \\
face & faz & cara & 0.41 \\
change & caer & cambia & 0.41 \\
concevoir & concebir & dise{\~n}ar & 0.18 \\
majeur & mayor & importante & 0.14 \\
\hline

\end{tabular}
\end{center}
\caption{\label{table:exp3}Extracted false friends for French-Spanish}
\end{table}

\subsubsection{Evaluation}

In this section we describe our overall results on identifying false friends for every language pair between English and five Romance languages: French, Italian, Spanish, Portuguese and Romanian.

\begin{table}[!ht]
\begin{center}
\begin{tabular}{l | l l l}
\hline
& Accuracy & Precision & Recall\\
\hline

Our method & 81.12 & 86.68 & 75.59 \\
(Castro et al) & 77.28 &  &  \\
WN Baseline & 69.57 & 85.82 & 54.50 \\
\hline

\end{tabular}
\end{center}
\caption{\label{table:exp1}Performance for Spanish-Portuguese using curated false friends test set}
\end{table}

We evaluate our method in two separate stages. First, we measure accuracy of false friend detection on a manually curated list of false friends and true cognates in Spanish and Portuguese, used in a previous study \cite{castro2018high}, and introduced in \cite{torres2011using}. This resource is composed by 710 Spanish-Portuguese word pairs: 338 true cognates and 372 false friends. We also compare our results to the ones reported in this study, which uses a method similar to ours (using a simple classifier that takes embedding similarities as features to identify false friends) and shows improvements over results in previous research. The results are show in Table \ref{table:exp1}. 

For the second part of the experiment, we use the list of cognates sets in English and Romance languages proposed by \cite{ciobanu_and_dinu_lrec} (the same that we used in our semantic divergence experiments), and try to automatically decide which of these are false friends.
Since manually built false friends lists are not available for every language pair that we experiment on, for the language pairs in this second experiment we build our gold standard by using a multilingual dictionary (WordNet) in order to infer false friends and true cognate relationships. We assume two cognates in different languages are true cognates if they occur together in any WordNet synset, and false friends otherwise.

\begin{table}[!ht]
\begin{center}
\small
\begin{tabular}{l | l l l}
\hline
& Accuracy & Precision & Recall\\
\hline

EN-ES & 76.58 & 63.88 & 88.46 \\
ES-IT & 75.80 & 41.66 & 54.05 \\
ES-PT & 82.10 & 40.0 & 42.85 \\
EN-FR & 77.09 & 57.89 & 94.28 \\
FR-IT & 74.16 & 32.81 & 65.62 \\
FR-ES & 73.03 & 33.89 & 69.96 \\
EN-IT & 73.07 & 33.76 & 83.87 \\
IT-PT & 76.14 & 29.16 & 43.75 \\
EN-PT & 77.25 & 59.81 & 86.48 \\
\hline

\end{tabular}
\end{center}
\caption{\label{table:exp4}Performance for all language pairs using WordNet as gold standard.}
\end{table}

We measure accuracy, precision, and recall, where:
\begin{itemize}
\item a \textit{true positive} is a cognate pair that are not synonyms in WordNet and are identified as false friends by the algorithm,
\item a \textit{true negative} is a pair which is identified as true cognates and is found in the same WordNet synset,
\item a \textit{false positive} is a word pair which is identified as a false friends pair by the algorithm but also appears as a synonym pair in WordNet,
\item and a \textit{false negative} is a pair of cognate words that are not synonyms in WordNet, but are also not identified as false friends by the algorithm.
\end{itemize}

We should also note that in the WordNet based method we can only evaluate results for only slightly over half of cognate pairs, since not all of them are found in WordNet. This also makes our corpus-based method more useful than a dictionary-based method, since it is able to cover most of the vocabulary of a language (given a large monolingual corpus to train embeddings on).

To be able to compare results to the ones evaluated on the manually built test set, we use the WordNet-based method as a baseline in the first experiment. Results for the second evaluation experiments are reported in Table \ref{table:exp4}. In this evaluation experiment we were able to measure performance for language pairs among all languages in our cognates set except for Romanian (which is not available in WordNet).

\section{Conclusions}

In this paper we proposed a method for computing the semantic divergence of cognates across languages. We relied on word embeddings and extended the pairwise metric to compute the semantic divergence across languages. Our results showed that Spanish and Portuguese are the closest languages, while Romanian is most dissimilar from Latin, possibly because it developed far from the Romance kernel. Furthermore, clustering the Romance languages based on the introduced semantic divergence measure results in a hierarchy that is consistent with the generally accepted tree of languages. When further including English in our experiments, we noticed that, even though most Latin words that entered English are probably borrowings (as opposed to inherited words), its similarity to Latin is close to that of the modern Romance languages. Our results shed some light on a new aspect of language similarity, from the point of view of cross-lingual semantic change.

We also proposed a method for detecting and possibly correcting false friends, and introduced a measure for quantifying the \textit{falseness} of a false friend, distinguishing between two categories: hard false friends and soft false friends. These analyses and algorithms for dealing with false friends can possibly provide useful tools for language learning or for (human or machine) translation.

In this paper we provided a simple method for detecting and suggesting corrections for false friends independently of context. There are, however, false friends pairs that are context-dependent - the cognates can be used interchangibly in some contexts, but not in others. In the future, the method using word embeddings could be extended to provide false friend correction suggestions in a certain context (possibly by using the word embedding model to predict the appropriate word in a given context).

\section*{Acknowledgements}
Research supported by BRD --- Groupe Societe Generale Data Science Research Fellowships. 

\bibliography{cognates}

\begin{thebibliography}{10}

\bibitem{artetxe2016learning}
Artetxe, M., Labaka, G., Agirre, E.:
\newblock Learning principled bilingual mappings of word embeddings while
  preserving monolingual invariance.
\newblock In: Proceedings of the 2016 Conference on Empirical Methods in
  Natural Language Processing. (2016)  2289--2294

\bibitem{bojanowski_et_al}
Bojanowski, P., Grave, E., Joulin, A., Mikolov, T.:
\newblock {Enriching Word Vectors with Subword Information}.
\newblock arXiv preprint arXiv:1607.04606 (2016)

\bibitem{campbell_1998}
Campbell, L.:
\newblock {Historical Linguistics. An Introduction}.
\newblock MIT Press (1998)

\bibitem{castro2018high}
Castro, S., Bonanata, J., Ros{\'a}, A.:
\newblock A high coverage method for automatic false friends detection for
  spanish and portuguese.
\newblock In: Proceedings of the Fifth Workshop on NLP for Similar Languages,
  Varieties and Dialects (VarDial 2018). (2018)  29--36

\bibitem{chen2016false}
Chen, Y., Skiena, S.:
\newblock False-friend detection and entity matching via unsupervised
  transliteration.
\newblock arXiv preprint arXiv:1611.06722 (2016)

\bibitem{ciobanu_and_dinu_lrec}
Ciobanu, A.M., Dinu, L.P.:
\newblock {Building a Dataset of Multilingual Cognates for the Romanian
  Lexicon}.
\newblock In: Proceedings of the Ninth International Conference on Language
  Resources and Evaluation, LREC 2014. (2014)  1038--1043

\bibitem{ciobanu_and_dinu_intelligibility}
Dinu, L.P., Ciobanu, A.M.:
\newblock {On the Romance Languages Mutual Intelligibility}.
\newblock In: Proceedings of the Ninth International Conference on Language
  Resources and Evaluation, {LREC} 2014. (2014)  3313--3318

\bibitem{eger_et_al}
Eger, S., Hoenen, A., Mehler, A.:
\newblock {Language Classification from Bilingual Word Embedding Graphs}.
\newblock In: Proceedings of COLING 2016, Technical Papers. (2016)  3507--3518

\bibitem{inkpen2005automatic}
Inkpen, D., Frunza, O., Kondrak, G.:
\newblock Automatic identification of cognates and false friends in french and
  english.
\newblock In: Proceedings of the International Conference Recent Advances in
  Natural Language Processing. Volume~9. (2005)  251--257

\bibitem{kondrak2003cognates}
Kondrak, G., Marcu, D., Knight, K.:
\newblock Cognates can improve statistical translation models.
\newblock In: Companion Volume of the Proceedings of HLT-NAACL 2003-Short
  Papers. (2003)

\bibitem{nakov2009unsupervised}
Nakov, S., Nakov, P., Paskaleva, E.:
\newblock Unsupervised extraction of false friends from parallel bi-texts using
  the web as a corpus.
\newblock In: Proceedings of the International Conference RANLP-2009. (2009)
  292--298

\bibitem{smith2017offline}
Smith, S.L., Turban, D.H., Hamblin, S., Hammerla, N.Y.:
\newblock Offline bilingual word vectors, orthogonal transformations and the
  inverted softmax.
\newblock arXiv preprint arXiv:1702.03859 (2017)

\bibitem{soegard_et_al}
S{\o}gaard, A., Goldberg, Y., Levy, O.:
\newblock {A Strong Baseline for Learning Cross-Lingual Word Embeddings from
  Sentence Alignments}.
\newblock In: {Proceedings of the 15th Conference of the European Chapter of
  the Association for Computational Linguistics, EACL 2017}. (2017)  765--774

\bibitem{st2017identifying}
St~Arnaud, A., Beck, D., Kondrak, G.:
\newblock Identifying cognate sets across dictionaries of related languages.
\newblock In: Proceedings of the 2017 Conference on Empirical Methods in
  Natural Language Processing. (2017)  2519--2528

\bibitem{torres2011using}
Torres, L.S., Alu{\'\i}sio, S.M.:
\newblock Using machine learning methods to avoid the pitfall of cognates and
  false friends in spanish-portuguese word pairs.
\newblock In: Proceedings of the 8th Brazilian Symposium in Information and
  Human Language Technology. (2011)

\bibitem{vulic_and_moens_2}
Vulic, I., Moens, M.:
\newblock {Cross-Lingual Semantic Similarity of Words as the Similarity of
  Their Semantic Word Responses}.
\newblock In: {Proceedings of Human Language Technologies: Conference of the
  North American Chapter of the Association of Computational Linguistics}.
  (2013)  106--116

\bibitem{vulic_and_moens}
Vulic, I., Moens, M.:
\newblock {Probabilistic Models of Cross-Lingual Semantic Similarity in Context
  Based on Latent Cross-Lingual Concepts Induced from Comparable Data}.
\newblock In: {Proceedings of the 2014 Conference on Empirical Methods in
  Natural Language Processing, EMNLP 2014}. (2014)  349--362

\end{thebibliography}
\bibliographystyle{splncs_srt}

\end{document}